\documentclass[11pt]{article}
\usepackage[margin=1in]{geometry}
\usepackage{times}
\usepackage{graphicx}
\usepackage{booktabs}
\usepackage{array}
\usepackage{float}
\usepackage{placeins}
\usepackage[table]{xcolor}
\usepackage{pifont}
\usepackage{enumitem}
\usepackage{fvextra}
\usepackage[round,authoryear]{natbib}
\setcitestyle{authoryear,round,citesep={;},aysep={,},yysep={;}}

\usepackage{amsmath,amsfonts,bm}

\def\eqref#1{equation~\ref{#1}}
\def\Eqref#1{Equation~\ref{#1}}

\def\1{\bm{1}}

\DeclareMathAlphabet{\mathsfit}{\encodingdefault}{\sfdefault}{m}{sl}
\SetMathAlphabet{\mathsfit}{bold}{\encodingdefault}{\sfdefault}{bx}{n}

\usepackage[hidelinks]{hyperref}
\usepackage{url}
\hypersetup{pdftitle={Beyond Dyadic Memory: Interaction-Aware Multimodal Memory with Adaptive Agentic Retrieval for Multi-Party Spoken Conversations},pdfauthor={Wenxu Jia, Xize Cheng, Zihan Zhang, Dongjie Fu, Linjun Li, Wenshi Chen, Yangyang Wu, Tao Jin}}
\title{Beyond Dyadic Memory: Interaction-Aware Multimodal Memory with Adaptive Agentic Retrieval for Multi-Party Spoken Conversations}
\author{Wenxu Jia$^{1,2,*}$, Xize Cheng$^{1,*}$, Zihan Zhang$^{1,*}$, Dongjie Fu$^1$\\
Linjun Li$^1$, Wenshi Chen$^2$, Yangyang Wu$^1$, Tao Jin$^{1,\dagger}$\\[0.6em]
\small $^1$Zhejiang University \qquad $^2$Meituan\\[0.3em]
\small $^*$Equal contribution (co-first authors). \quad $^\dagger$Corresponding author.}
\date{}

\begin{document}

\maketitle

\begin{abstract}
Long-term memory enables agents to accumulate information and reason across sessions, yet existing research primarily focuses on dyadic text or image-text conversations, leaving long-term memory for multi-party spoken conversations underexplored. This setting requires preserving conversational content, identifying participants across sessions, and retaining who speaks to whom. To this end, we propose VoxPolyMem, an interaction-aware multimodal memory framework combining incremental speaker identification with a memory hierarchy comprising interaction memory, fact memory, and participant profiles. We formulate retrieval as sequential decision-making, where an agent rewrites queries and selects retrieval tools and memory layers based on accumulated evidence to address information gaps. We further introduce Evidence-Gain GRPO (EG-GRPO), which uses round-wise credit assignment to encourage complementary evidence acquisition. We also construct VoxPolyBench to evaluate memory evolution, personalized answering, memory retrieval and reasoning, and interaction reasoning and attribution in multi-party spoken conversations. VoxPolyMem achieves an overall score of 85.0 on VoxPolyBench, surpassing the strongest evaluated baseline by 23.6 points. On Mem-Gallery and H2HMem-Multi, it scores 89.6 and 74.4, respectively, exceeding the strongest evaluated public memory baselines by over 8 points each. These results highlight its potential for persistent, personalized assistance in multi-party multimodal interactions. Code and datasets are available at \url{https://voxpolymem.github.io/VoxPolyBench/demo/}.
\end{abstract}

\section{Introduction}

Long-term memory enables agents to retain information and reason across sessions~\citep{memory_survey,memory_survey_v2,memory_survey_v3}. Recent systems consolidate conversational facts, link related memories, and incorporate multimodal information~\citep{chhikara2025mem0,xu2025amem,liu2025memverse,feng2026m2a}. However, evaluations largely focus on dyadic text or image-text conversations~\citep{wu2024longmemeval,bei2026memgallery}. In-car, meeting, and household assistants often engage with multiple participants through speech, requiring them to remember information and interactions across sessions. Recent work extends memory evaluation to multi-party text or image-text histories~\citep{yang2026groupmembench} and explores multi-speaker speech understanding and voice-based personalization~\citep{xiong2026msibench,alratrout2026afa}. Yet long-term memory for multi-party spoken conversations remains underexplored, with limited benchmarks for evaluating recurring speaker identification, participant-specific memories, and evolving interactions across sessions.


Constructing memory for these conversations requires more than preserving transcribed content~\citep{inoue2025addressee}. It also requires extracting voice representations from audio and linking them to participant identities across sessions. Storing audio solely as ASR transcripts leaves these acoustic identity cues unmodeled. Moreover, existing memory methods organize conversational histories through temporal graphs and semantic abstraction~\citep{rasmussen2025zep,liu2026simplemem}, but do not explicitly store interaction roles, such as who made a statement and to whom it was addressed. Preserving speaker identity and interaction roles supports the extraction of factual memories and participant profiles that capture individual preferences, commitments, and relationships. Retrieving supporting evidence poses a further challenge, as information may span sessions, memory layers, and modalities, with tools ranging from keyword search to cross-modal retrieval~\citep{yeo2026universalrag}. Fixed retrieval strategies may overlook information needs that emerge as evidence accumulates. Iterative retrieval enables multi-step evidence gathering~\citep{trivedi2023ircot}, while learned search optimizes retrieval using terminal answer rewards~\citep{jin2025searchr1}. However, terminal rewards do not explicitly isolate the contribution of each retrieval round to acquiring new supporting evidence.


To address these challenges, we propose VoxPolyMem, an interaction-aware multimodal memory framework that combines incremental speaker identification with interaction memory, fact memory, and participant profiles. A directed interaction graph records who said what to whom. A dialogue window provides recent conversational context for extracting self-contained facts and participant attributes, while profiles link participant names and aliases to their voice representations. Building on this hierarchy, a trainable retrieval agent selects memory layers and tools and rewrites queries based on accumulated evidence and previous actions. We train the agent with Evidence-Gain GRPO (EG-GRPO), an adaptation of Group Relative Policy Optimization (GRPO)~\citep{shao2024deepseekmath}. At each retrieval state, EG-GRPO compares candidate actions through the coverage and ranking of newly acquired supporting evidence retained in the context. Previously retained evidence receives no additional credit, encouraging complementary retrieval across memory layers and modalities within a fixed budget.

To evaluate long-term memory in multi-party spoken conversations, we construct VoxPolyBench, comprising 1,527 QA pairs over 176 sessions and 18.9 hours of synthesized speech across 18 scenarios. Each scenario follows recurring participants across sessions, enabling evaluation of cross-session participant identification, memory evolution, personalized answering, retrieval and reasoning, and interaction attribution. Our contributions are threefold:
\begin{itemize}[left=0em, itemsep=-0.5pt, label=\textbullet]
    \item We introduce VoxPolyBench, a benchmark for long-term memory in multi-party spoken conversations, covering recurring participant identification, memory evolution, personalized answering, retrieval and reasoning, and interaction attribution.
    \item We propose VoxPolyMem, integrating incremental speaker identification, interaction-aware hierarchical memory, and adaptive retrieval. Its retrieval agent selects memory layers and tools and rewrites queries, with EG-GRPO rewarding newly acquired supporting evidence across retrieval rounds.
    \item Experiments on VoxPolyBench, Mem-Gallery, and H2HMem-Multi demonstrate consistent improvements in overall answer quality. Ablations highlight the benefits of interaction-aware memory, while retrieval-policy comparisons show improved evidence recall with fewer retrieval rounds.
\end{itemize}

\section{Related Work}

\subsection{Long-Term Memory Construction and Retrieval}

Agent memory systems organize information from prior conversations and observations for subsequent retrieval and reasoning. Mem0 consolidates facts, A-Mem links contextual notes, and Zep maintains temporal knowledge graphs~\citep{chhikara2025mem0,xu2025amem,rasmussen2025zep}. LightMem and SimpleMem improve efficiency through staged consolidation and semantic compression~\citep{fang2025lightmem,liu2026simplemem}. Multimodal systems introduce hierarchical, episodic, and semantic memories for image-text and audiovisual experiences~\citep{liu2025memverse,feng2026m2a,lin2025hippomm,long2025seeing,lian2026mmmem}. AFA uses voice-based identification to separate users' memories~\citep{alratrout2026afa}. For shared multi-party conversations, VoxPolyMem links recurring speaker identities to directed interaction relations, preserving who said what to whom when extracting facts and participant profiles.

Retrieval methods combine routing, reasoning, and policy learning. UniversalRAG routes queries across modalities and granularities, while IRCoT interleaves retrieval with reasoning~\citep{yeo2026universalrag,trivedi2023ircot}. Search-R1 and Memory-R1 use outcome rewards to optimize search or memory operations~\citep{jin2025searchr1,yan2025memoryr1}. LeTS combines process and outcome rewards, while Mem-T provides dense credit for memory operations~\citep{zhang2025lets,yue2026memt}. InfoReasoner estimates retrieval gains through reductions in semantic uncertainty~\citep{hu2026inforeasoner}. EG-GRPO rewards the coverage and ranking of newly acquired supporting evidence retained after each action. Rewards are computed after context selection, with previously retained evidence excluded from both terms.

\subsection{Long-Term Memory and Spoken Interaction Benchmarks}

LoCoMo and LongMemEval evaluate long-term conversational recall and reasoning~\citep{maharana2024locomo,wu2024longmemeval}. Mem-Gallery extends evaluation to image-text histories, while H2HMem, GroupMemBench, and EverMemBench cover multi-party memory settings~\citep{bei2026memgallery,zhu2026h2hmem,yang2026groupmembench,hu2026evermembench}. These settings do not jointly evaluate acoustic identity recovery and participant-dependent memory. For spoken interaction, ContextDialog tests conversational recall, MSU-Bench evaluates multi-speaker understanding, and Vox-Infinity studies long-context speech understanding~\citep{kim2025does,wang2025msubench,xu2026voxinfinity}. VoxPolyBench jointly evaluates recurring speaker identification, memory evolution, personalized answering, retrieval and reasoning, and speaker and addressee attribution over multi-session, multi-party spoken histories.

\section{VoxPolyBench}
\label{sec:benchmark}

VoxPolyBench comprises 18 scenarios across domains including telemarketing, meetings, in-car assistance, and household assistance. Each scenario contains a dialogue history spanning 8--12 sessions with recurring participants. In total, the benchmark contains 176 sessions, 18.9 hours of synthesized speech, and 1,527 QA pairs. The QA pairs cover memory evolution, personalized answering, retrieval and reasoning, and interaction attribution. Table~\ref{tab:benchmark-capability-comparison} compares its scope with existing benchmarks, and Appendix~\ref{app:dataset-composition} provides detailed statistics.

\begin{figure}[t]
    \centering
    \setlength{\abovecaptionskip}{5pt}
    \includegraphics[width=\linewidth]{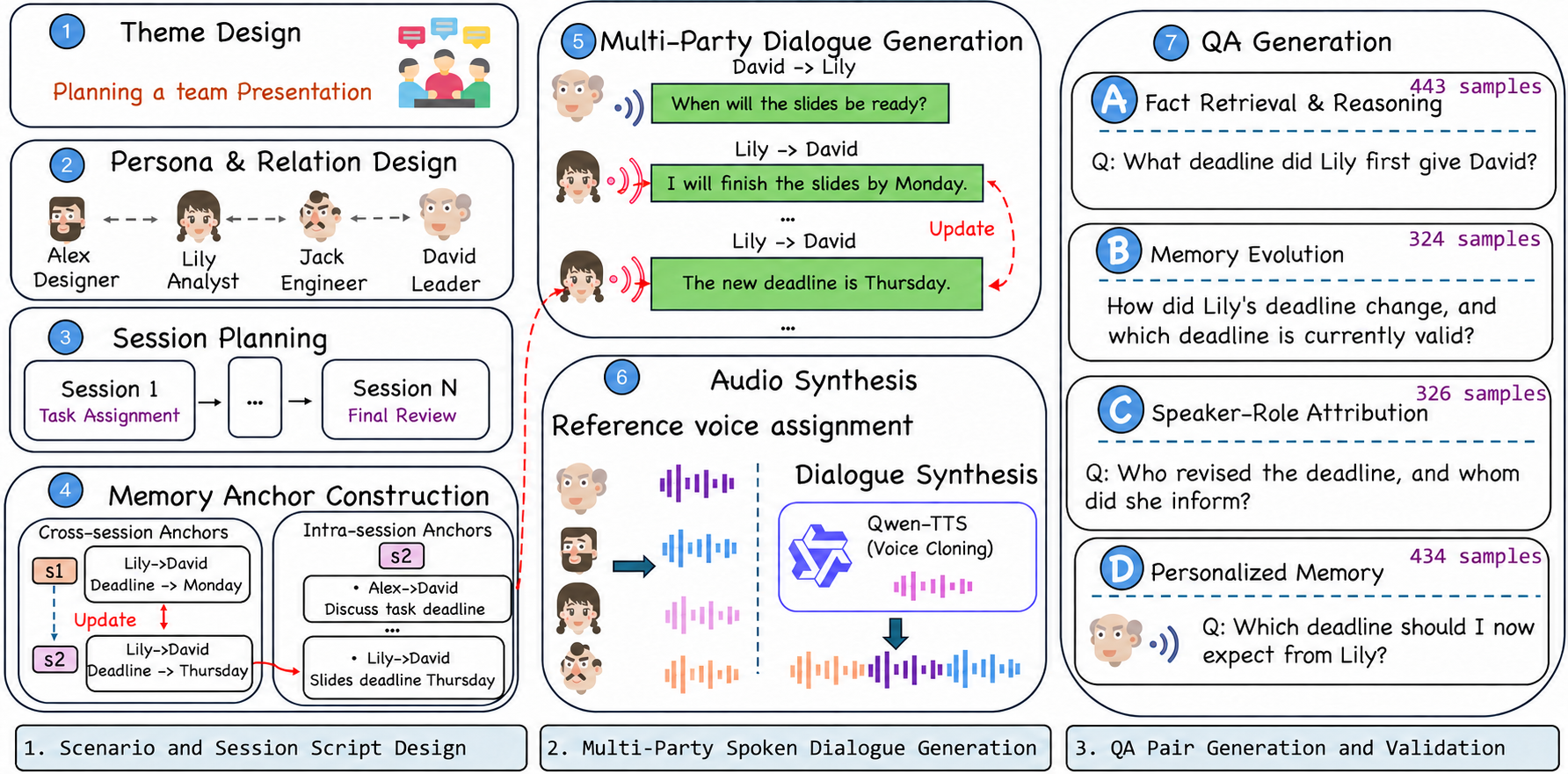}
    \caption{Construction of VoxPolyBench. Scenario plans and event anchors guide multi-party dialogue and speech generation, followed by QA construction and validation. Task counts follow the dataset taxonomy in Appendix~\ref{app:dataset-composition}; result aggregation is specified in Appendix~\ref{app:metrics}.}
    \label{fig:data-construction}
    \vspace{-0.5cm}
\end{figure}

We construct the benchmark in three stages, using event anchors to plan how information and participant interactions develop across sessions. Figure~\ref{fig:data-construction} summarizes the pipeline.

\noindent\textbf{Stage 1: Hierarchical Scenario and Event Planning.}
We first define each scenario's theme and participant settings, including identities, roles, long-term attributes, and interpersonal relationships. We then plan session topics and event progression, specifying event anchors within and across sessions. Within-session anchors describe local events and participant interactions, while cross-session anchors track the persistence, updates, and conflicts of facts and preferences. Each anchor records the relevant participants, time, and event content; interaction anchors additionally specify speakers and addressees. Human reviewers check each session's anchors before dialogue generation and revise any identified errors or implausible event transitions to maintain logical consistency and support natural dialogue generation.

\noindent\textbf{Stage 2: Multi-Party Dialogue and Speech Generation.}
We use GPT-4.1 to generate multi-party dialogues from session topics and event anchors, conditioned on participant settings, prior event states, and subsequent anchors to maintain continuity across sessions. Speech synthesis uses a fixed reference voice for each participant across sessions, with telephone numbers and alphanumeric identifiers rendered for character-by-character pronunciation. We check synthesis duration and use ASR to identify transcription mismatches, identifier errors, and suspected repetitions. Targeted resynthesis, transcription rechecks, and manual spot checks address identified quality issues. Final evaluation with Whisper large-v3-turbo over all dialogue audio segments yields a corpus-level word error rate (WER) of 2.14\% against normalized source texts. Appendix~\ref{app:quality-control} details the quality-control and evaluation procedures.

\noindent\textbf{Stage 3: QA Generation and Validation.}
We construct QA pairs from event anchors and the generated dialogues, designing questions around event states and their changes, determining answers from dialogue content, and annotating the dialogue turns that support each answer. Questions examine both the integration and updating of memories across sessions and whom the information concerns, who expressed it, and to whom it was addressed. Human reviewers and an LLM check event consistency, participant and interaction assignments, question clarity, and evidence sufficiency. Correctable errors are revised and unsupported or ambiguous QA pairs removed, following Appendix~\ref{app:quality-control}.

\begingroup
\definecolor{VPGreen}{HTML}{237A57}
\definecolor{VPAmber}{HTML}{B7791F}
\definecolor{VPGray}{HTML}{777777}
\definecolor{VPRow}{HTML}{EEF7F3}
\newcommand{\vpyes}{\textcolor{VPGreen}{\ding{51}}}
\newcommand{\vppart}{\textcolor{VPAmber}{$\circ$}}
\newcommand{\vpno}{\textcolor{VPGray}{--}}

\begin{table}[htbp]
    \centering
    \caption{Comparison of spoken-dialogue and memory benchmarks. \vpyes\ denotes explicit evaluation, \vppart\ partial or adjacent coverage, and \vpno\ no targeted evaluation. Interaction attribution covers both speakers and addressees; personalization requires answers conditioned on the querying participant. Sources and comparison criteria are detailed in Appendix~\ref{app:benchmark-comparison}.}
    \label{tab:benchmark-capability-comparison}
    \small
    \setlength{\tabcolsep}{2pt}
    \renewcommand{\arraystretch}{1.18}
    \begin{tabular*}{\linewidth}{@{\extracolsep{\fill}}llcccccc@{}}
        \toprule
        \textbf{Benchmark} & \textbf{Modality} & \shortstack{Multi-\\party} & \shortstack{Multi-\\session} & \shortstack{Cross-session\\identity} & \shortstack{Interaction\\attribution} & \shortstack{Personali-\\zation} & \shortstack{Memory\\update} \\
        \midrule
        ContextDialog & Speech & \vpno & \vpno & \vpno & \vpno & \vpno & \vpno \\
        MSU-Bench & Speech & \vpyes & \vpno & \vpno & \vppart & \vpno & \vpno \\
        AFA & Speech & \vpno & \vppart & \vpyes & \vppart & \vpyes & \vppart \\
        Vox-Infinity & Speech & \vpno & \vpno & \vpno & \vpno & \vpno & \vpno \\
        Mem-Gallery & Text+Image & \vpno & \vpyes & \vpno & \vpno & \vpno & \vpyes \\
        H2HMem & Text+Image & \vpyes & \vpyes & \vpyes & \vpno & \vpno & \vpyes \\
        \rowcolor{VPRow}
        \textbf{VoxPolyBench} & Speech & \vpyes & \vpyes & \vpyes & \vpyes & \vpyes & \vpyes \\
        \bottomrule
    \end{tabular*}
    \vspace{-0.5cm}
\end{table}
\endgroup

\section{VoxPolyMem}
\label{sec:method}

\begin{figure}[t]
    \centering
    \setlength{\abovecaptionskip}{5pt}
    \includegraphics[width=\linewidth]{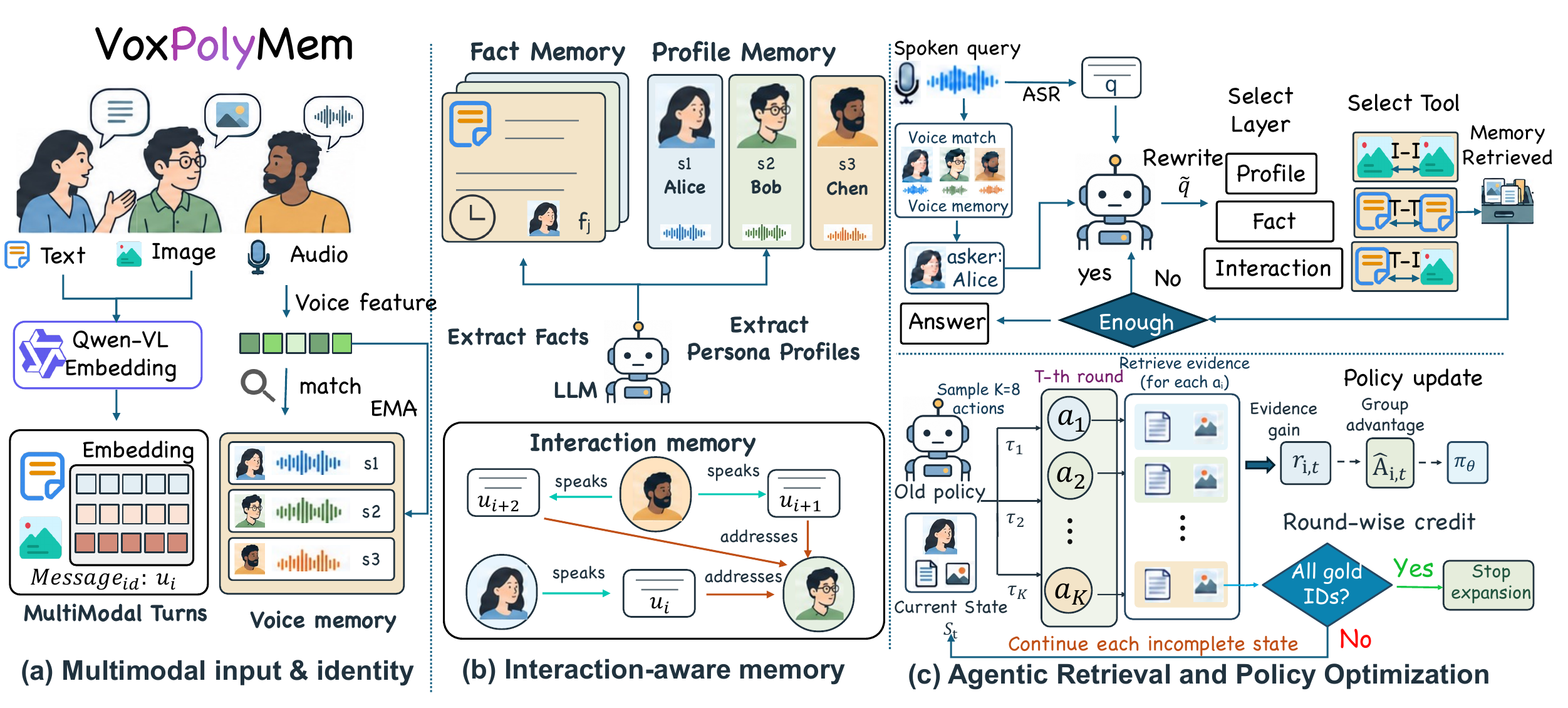}
    \caption{VoxPolyMem identifies recurring speakers, constructs interaction memory, facts, and participant profiles, and adaptively retrieves evidence across memory layers and modalities.}
    \label{fig:method-overview}
    \vspace{-0.5cm}
\end{figure}

\noindent\textbf{Overview.}
\label{sec:method-overview}
As illustrated in Figure~\ref{fig:method-overview}, VoxPolyMem connects input processing, memory construction, and query answering. For incoming multimodal conversations, online speaker identification assigns speaker identifiers to audio turns in Section~\ref{sec:speaker-grounding}. These identifiers and conversational content support the construction of interaction memory and the joint extraction of facts and participant profiles in Section~\ref{sec:hierarchical-memory}. Given a query, the retrieval agent uses profile voice memory to identify spoken-query participants, then iteratively rewrites queries and selects memory layers and tools to gather supporting evidence for answering in Section~\ref{sec:adaptive-retrieval}. EG-GRPO trains this policy to acquire complementary evidence across retrieval rounds.

\subsection{Online Speaker Identification}
\label{sec:speaker-grounding}

To distinguish recurring participants without a predefined speaker count, we maintain an online speaker memory. Acoustic matching assigns anonymous identifiers, and accumulated acoustic evidence reconciles fragmented identifiers. Names and aliases are associated later during LLM-based memory extraction in Section~\ref{sec:hierarchical-memory}.

\noindent\textbf{Speaker encoding.}
For audio $x_i^{\mathrm{aud}}$ at turn $i$, an ECAPA-TDNN encoder $E_{\mathrm{spk}}$~\citep{desplanques2020ecapa} produces the voiceprint embedding $\mathbf e_i$ in \eqref{eq:speaker-matching}. The operator $\operatorname{norm}$ applies $L_2$ normalization, so inner products measure cosine similarity.

\noindent\textbf{Incremental identity matching.}
We maintain a set $\mathcal V$ of entries $(k,\mathbf v_k)$, where $k$ is an anonymous \texttt{speaker\_id} and $\mathbf v_k$ its stored, normalized voiceprint embedding. The set is initially empty, so the first spoken turn creates an identifier with $\mathbf e_i$ as its voiceprint embedding. For each subsequent audio turn, we compare $\mathbf e_i$ with all stored embeddings using cosine similarity. \Eqref{eq:speaker-matching} selects the most similar entry, yielding identifier $k^*$ and similarity score $s_i$. If $s_i\geq\tau_{\mathrm{match}}$, we assign the turn to $k^*$. Otherwise, no stored speaker provides a sufficiently close match, so we create a new identifier initialized with $\mathbf e_i$. This procedure accommodates newly appearing speakers without requiring their number in advance. Non-audio turns leave $\mathcal V$ unchanged.
\begin{equation}
    \mathbf e_i=\operatorname{norm}\!\left(E_{\mathrm{spk}}(x_i^{\mathrm{aud}})\right),
    \quad k^*=\underset{k:\,(k,\mathbf v_k)\in\mathcal V}{\arg\max}\;\mathbf e_i^\top\mathbf v_k,
    \qquad s_i=\mathbf e_i^\top\mathbf v_{k^*}.
    \label{eq:speaker-matching}
\end{equation}

\noindent\textbf{Confidence-gated memory updates.}
To accommodate acoustic variation without reinforcing uncertain matches, we update a matched embedding by exponential moving average (EMA) only when $s_i\geq\tau_{\mathrm{update}}>\tau_{\mathrm{match}}$. The update in \eqref{eq:speaker-memory-update} uses weight $\alpha$ for the current observation. Scores between the thresholds associate the turn with an existing identifier but leave its embedding unchanged.
\begin{equation}
    \mathbf v_{k^*}\leftarrow
    \operatorname{norm}\!\left((1-\alpha)\mathbf v_{k^*}+\alpha\mathbf e_i\right).
    \label{eq:speaker-memory-update}
\end{equation}

Acoustic variation may cause the same speaker to receive multiple provisional speaker IDs. After each session, we revisit these IDs using accumulated voiceprint evidence and conservatively merge those supported as belonging to the same speaker, while retaining their underlying voiceprint embeddings. This reduces identity fragmentation without requiring a participant roster or target speaker count. Appendix~\ref{app:speaker-identification} provides the merging criteria and configuration.

\subsection{Hierarchical Interaction-Aware Multimodal Memory}
\label{sec:hierarchical-memory}

We organize conversational information into interaction memory $\mathcal M_{\mathrm{int}}$, fact memory $\mathcal M_{\mathrm{fact}}$, and participant profiles $\mathcal M_{\mathrm{prof}}$. Interaction memory preserves multimodal messages and their participant relations as source evidence. Fact memory captures self-contained statements grounded in these interactions, while participant profiles organize identity, background, and preferences around individual participants. Facts and profiles are jointly extracted from local conversational context.

\noindent\textbf{Multimodal interaction memory.}
Each message retains its text, original multimodal attachments, timestamp, and session position. Text, image captions, and automatic speech recognition (ASR) transcripts from \texttt{whisper-turbo} support semantic extraction. Qwen3-VL-Embedding~\citep{li2026qwen3vlembedding} encodes text and images for content retrieval, separately from the voiceprint embeddings used for speaker identification. We represent interaction memory as a directed graph $\mathcal M_{\mathrm{int}}=(\mathcal P\cup\mathcal U,\mathcal E)$, where $\mathcal P$ and $\mathcal U$ contain participant and message nodes. The edge set records who speaks and to whom:
\begin{equation}
    \mathcal E=\bigcup_i\left(
    \left\{\mathrm{speaker}_i\xrightarrow{\mathrm{speaks}}u_i\right\}
    \cup
    \left\{u_i\xrightarrow{\mathrm{addresses}}p\;\middle|\;p\in\mathcal A_i\right\}
    \right).
    \label{eq:interaction-edges}
\end{equation}
For message $u_i$, $\mathrm{speaker}_i$ denotes its speaker and $\mathcal A_i$ its addressee set. Addressees are inferred through the context-aware extraction described below. Unknown addressees introduce no participant-specific edge.

\noindent\textbf{Context-aware fact and participant memory.}
An LLM processes the current turn and up to three preceding turns to associate anonymous \texttt{speaker\_id}s with participant names and aliases while jointly extracting facts, participant attributes, addressees, and temporal references. Identifiers resolved to the same participant are linked through an identity mapping while retaining their voiceprint embeddings. The set $\mathcal K_p$ contains the identifiers associated with participant $p$, whose profile also records background and preferences. Fact memory stores self-contained statements with their speakers, addressees, and time. The records in \eqref{eq:identity-grounded-memory} summarize these fields; image-related facts also retain their image identifiers. Each fact and profile retains \texttt{refer\_id} links to its source dialogue turns.
\begin{equation}
    \begin{aligned}
        \mathcal M_{\mathrm{fact}}
        &= \left\{\left(\mathrm{content}_j,\,\mathrm{speaker}_j,\,
            \mathrm{addressee}_j,\,\mathrm{time}_j\right)\right\}_j,\\[4pt]
        \mathcal M_{\mathrm{prof}}
        &= \left\{\left(\mathrm{name}_p,\,\mathrm{aliases}_p,\,
            \{(k,\mathbf v_k)\}_{k\in\mathcal K_p},\,
            \mathrm{attributes}_p\right)\right\}_{p\in\mathcal P}.
    \end{aligned}
    \label{eq:identity-grounded-memory}
\end{equation}
The index $j$ ranges over extracted facts, whose fields record content, speaker, addressees, and time. Each participant profile stores a name, aliases, associated voiceprint embeddings, and attributes comprising background and preferences.

\subsection{Trainable Agentic Memory Retrieval}
\label{sec:adaptive-retrieval}

\noindent\textbf{Evidence-conditioned retrieval.}
Queries may require evidence across memory layers and modalities, with intermediate results guiding subsequent retrieval. For spoken queries, voiceprint matching against profiles identifies $\mathrm{asker}_q$; unmatched speakers remain unknown. The state and action are

\begin{equation}
    \begin{aligned}
        s_t &= (q,\mathrm{asker}_q,C_{t-1},H_{t-1},b_t),\\
        a_t &= (\ell_t,\rho_t,\widetilde q_t,f_t)
        \sim\pi_\theta(\cdot\mid s_t),
        \qquad \rho_t\in\mathcal{R}(\ell_t).
    \end{aligned}
    \label{eq:agent-action}
\end{equation}
Here, $t$ indexes rounds. The state contains query $q$, asker identity, retained evidence $C_{t-1}$, action history $H_{t-1}$, and remaining budget $b_t$. Policy $\pi_\theta$ selects layer $\ell_t$ (interaction, fact, or profile), tool $\rho_t\in\mathcal R(\ell_t)$, rewritten query $\widetilde q_t$, and optional speaker/addressee filters $f_t$. Tools support text-to-text retrieval through vector similarity or BM25, text-to-image retrieval through image descriptions, and image-to-image retrieval through image embeddings. Reciprocal rank fusion (RRF), deduplication, and context selection update the retained evidence. An LLM identifies missing information to guide further retrieval, or answers when evidence is sufficient or three rounds are exhausted.

\noindent\textbf{Evidence-Gain GRPO.}
\label{sec:evidence-gain}
To support adaptive retrieval, EG-GRPO adapts GRPO~\citep{shao2024deepseekmath} to reward newly acquired supporting evidence retained in context. At each state, $\pi_{\mathrm{old}}$ samples $K=8$ independent candidates sharing the pre-action context but not retrieved results. They form one GRPO group, indexed by $i$ below.

After RRF fusion and deduplication, candidate $i$ retains up to $k=15$ memory records in $C_i$, including fact and profile contents. For reward calculation, interaction-turn IDs and \texttt{refer\_id} links define the source-ID set $U_i$. Let $G_q\ne\emptyset$ contain annotated supporting IDs and $B$ IDs represented in earlier retained contexts on this branch, initially empty. Then $D_i=(U_i\cap G_q)\setminus B$ contains newly retained supporting IDs. For $e\in U_i$, $p_i(e)$ is the earliest rank of a memory in $C_i$ linked to $e$.
\begin{equation}
    \begin{aligned}
        N_i &= \frac{1}{Z_q}\sum_{e\in D_i}\frac{1}{\log_2(p_i(e)+1)},
        \qquad Z_q=\sum_{j=1}^{\min(k,|G_q|)}\frac{1}{\log_2(j+1)},\\
        r_i &= \lambda\frac{|D_i|}{|G_q|}+(1-\lambda)N_i,
        \qquad \widehat A_i=\frac{r_i-\mu}{\sigma+\delta}.
    \end{aligned}
    \label{eq:evidence-objective}
\end{equation}
In \eqref{eq:evidence-objective}, coverage rewards newly retained supporting IDs, while $N_i$ discounts them by memory rank using fixed scale $Z_q$. Each source ID contributes once, even if linked to multiple records. The weight $\lambda\in[0,1]$ balances coverage and ranking. The advantage $\widehat A_i$ uses the mean $\mu$ and standard deviation $\sigma$ of this state's rewards, with $\delta>0$ for stability.

Each candidate creates a successor $s_{t+1}^{(i)}$ with its selected evidence $C_i$, updated action history, and budget $b_{t+1}^{(i)}=b_t-1$. During training, we expand
\begin{equation}
    \mathcal S_{\mathrm{next}}(s_t)=\left\{s_{t+1}^{(i)}\;\middle|\;
        G_q\setminus U_i\ne\emptyset,\quad
        b_{t+1}^{(i)}>0\right\}.
    \label{eq:continuing-states}
\end{equation}
Each successor in $\mathcal S_{\mathrm{next}}$ independently samples another $K$ actions. Unexpanded successors' actions remain in their current group. All groups update the shared policy through \eqref{eq:grpo-objective}.

\begin{equation}
    \mathcal J_{\mathrm{GRPO}}(\theta)
    =\mathbb E\Bigl[
    \tfrac1K\sum_{i=1}^{K}\tfrac1{|a_i|}\sum_{n=1}^{|a_i|}
    \bigl\{\min\!\bigl[
        w_{i,n}\widehat A_i,\,
        \operatorname{clip}(w_{i,n},1-\epsilon,1+\epsilon)
        \widehat A_i
    \bigr]-\beta\mathrm{KL}_{i,n}\bigr\}\Bigr].
    \label{eq:grpo-objective}
\end{equation}
Here, $n$ indexes the $|a_i|$ action tokens, distinct from memory positions $p_i(e)$. The ratio $w_{i,n}$ compares current and old policies under the same state and action prefix; $\mathrm{KL}_{i,n}$ measures divergence from a fixed reference policy. The coefficients $\epsilon$ and $\beta$ control clipping and KL regularization. Advantages update only generated layer, tool, query, and filter tokens. Input, retrieved, and answer tokens are excluded; the answer model remains fixed. Gold annotations are training-only. Appendix~\ref{app:eg-grpo} details source ordering, history updates, and optimization.

\section{Experiments}
\label{sec:experiments}

We evaluate VoxPolyMem on multi-party spoken and public image-text memory benchmarks, examining overall performance, component contributions, and retrieval policy optimization. Beyond synthesized speech in VoxPolyBench, the speaker tracker achieves attribution accuracies of 95.0\% on IEMOCAP and 87.9\% on AMI recordings with reference utterance boundaries (Appendix~\ref{app:speaker-ablation}).

\subsection{Experimental Setup}
\label{sec:experimental-setup}

\noindent\textbf{Evaluation setup.}
We compare VoxPolyMem with ten text-based and multimodal baselines on the held-out test splits of VoxPolyBench, Mem-Gallery~\citep{bei2026memgallery}, and H2HMem-Multi~\citep{zhu2026h2hmem}. On VoxPolyBench, all methods construct semantic memory from the same \texttt{whisper-turbo} ASR transcripts rather than the original dialogue scripts. VoxPolyMem and its ablations perform speaker identification automatically, while baselines receive ground-truth speaker identification labels. The public benchmarks retain their native text and visual inputs. All methods and ablations use GPT-4.1-mini as the answer model, with shared evaluation questions and rubric and a final-context budget of $k=15$ retrieved records within each benchmark. GPT-4.1-mini and GPT-4.1 independently judge each answer. We average their scores per answer and then across test questions to obtain Score. Recall measures the fraction of annotated supporting dialogue turns represented in the final context, averaged over questions with nonempty annotations. It is reported on VoxPolyBench and Mem-Gallery, since H2HMem-Multi lacks these annotations. Both metrics are multiplied by 100 and rounded to one decimal. Appendix~\ref{app:evaluation-details} details baselines, metrics, and evaluation protocols, while Appendix~\ref{app:human-llm-judge-agreement} reports the human audit.

\noindent\textbf{RL training.}
We train an 8B Qwen-VL retrieval policy for one epoch with EG-GRPO. GPT-4.1-mini handles memory extraction. Training uses approximately 900 QA instances: 300 from Mem-Gallery, 400 from LoCoMo~\citep{maharana2024locomo}, and 200 from VoxPolyBench. These datasets annotate supporting dialogue turns for each question, providing supervision for EG-GRPO. Table~\ref{tab:train-test-counts} reports evaluation split counts. The annotations determine rewards and training-time stopping based on current evidence coverage and are unavailable at inference.

\subsection{Main Results}
\label{sec:voxpolybench-results}
\label{sec:external-generalization}

\begin{table}[!t]
    \centering
    \small
    \renewcommand{\arraystretch}{1.12}
    \caption{Main results across three memory benchmarks ($\uparrow$). For VoxPolyBench, RR/EC/P/IA denote the retrieval and reasoning, evolution and conflict, persona, and interaction attribution subsets, respectively. For Mem-Gallery, E\&A/R/KM denote the extraction and adaptation, reasoning, and knowledge management subsets, respectively. For H2HMem-Multi, A/R/Rec. denote the memory application, memory reasoning, and memory recall subsets, respectively. Aver denotes the mean score within each benchmark. Average weights benchmark scores by their test-question counts. Bold and underlined values mark the best and second-best results.}
    \label{tab:voxpolybench-main}
    \label{tab:external-generalization}
    {\footnotesize
    \setlength{\tabcolsep}{0.8pt}
    \begin{tabular*}{\linewidth}{@{\extracolsep{\fill}}lrrrrrrrrrrrrrr@{}}
        \toprule
        & \multicolumn{5}{c}{\textbf{VoxPolyBench}} & \multicolumn{4}{c}{\textbf{Mem-Gallery}} & \multicolumn{4}{c}{\textbf{H2HMem-Multi}} & \multicolumn{1}{c}{\textbf{Average}} \\
        \cmidrule(lr){2-6}\cmidrule(lr){7-10}\cmidrule(lr){11-14}\cmidrule(lr){15-15}
        Method & RR & EC & P & IA & \textbf{Aver} & E\&A & R & KM & \textbf{Aver} & A & R & Rec. & \textbf{Aver} & \textbf{Score} \\
        \midrule
        mem0~\citeyearpar{chhikara2025mem0} & 79.1 & 79.6 & 40.5 & 42.6 & 60.4 & 50.6 & 61.2 & 81.9 & 60.6 & 71.3 & 29.8 & 53.9 & 54.6 & 60.2 \\
        A-Mem~\citeyearpar{xu2025amem} & 76.1 & 77.3 & 34.8 & 32.4 & 55.3 & 42.7 & 55.5 & 72.5 & 53.0 & 70.0 & 35.7 & 48.5 & 54.1 & 54.1 \\
        LightMem~\citeyearpar{fang2025lightmem} & 83.5 & 80.2 & 37.0 & 38.1 & 59.9 & 40.2 & 41.8 & 79.0 & 49.4 & 67.7 & 32.7 & 59.3 & 55.4 & 54.5 \\
        \midrule
        AUGUSTUS~\citeyearpar{jain2025augustus} & 77.1 & 78.9 & 38.1 & 32.2 & 56.8 & 77.1 & 68.0 & 74.0 & 73.8 & 77.3 & 48.2 & 66.5 & 66.0 & 65.6 \\
        UniversalRAG~\citeyearpar{yeo2026universalrag} & 77.8 & 80.1 & 39.6 & 35.1 & 58.3 & 77.1 & 80.7 & 75.7 & 77.8 & 62.0 & \textbf{58.3} & 72.5 & 64.0 & 68.1 \\
        M2A~\citeyearpar{feng2026m2a} & 82.7 & 80.2 & 37.6 & 45.0 & 61.3 & 49.3 & 38.2 & 77.2 & 52.4 & 56.0 & 52.6 & \textbf{77.0} & 61.1 & 57.0 \\
        MemVerse~\citeyearpar{liu2025memverse} & 84.7 & 87.3 & 37.7 & 32.3 & 60.7 & 71.2 & 63.2 & 75.7 & 69.9 & 67.0 & 39.4 & 64.9 & 58.7 & 65.0 \\
        NGMemory~\citeyearpar{fisher2025ngm} & 62.9 & 66.5 & 40.2 & 44.3 & 53.3 & 75.5 & 62.4 & 78.8 & 72.5 & 81.3 & 24.7 & 35.3 & 52.0 & 62.5 \\
        HippoMM~\citeyearpar{lin2025hippomm} & 62.7 & 47.5 & 34.5 & 61.3 & 51.2 & 46.4 & 67.1 & 79.0 & 59.6 & 63.3 & 21.9 & 42.2 & 45.6 & 54.9 \\
        M3-Agent~\citeyearpar{long2025seeing} & 70.5 & 68.4 & 25.2 & 40.0 & 50.7 & 68.8 & 55.8 & 81.1 & 67.9 & 49.7 & 3.1 & 18.1 & 27.4 & 57.6 \\
        \midrule
        w/o RL & \underline{85.5} & \underline{90.8} & \underline{75.9} & \underline{86.0} & \underline{84.0} & \underline{87.9} & \underline{83.9} & \underline{85.9} & \underline{86.3} & \underline{85.0} & 52.0 & 72.1 & \underline{72.1} & \underline{84.4} \\
        \textbf{VoxPolyMem} & \textbf{86.4} & \textbf{92.1} & \textbf{76.1} & \textbf{87.5} & \textbf{85.0} & \textbf{91.4} & \textbf{87.3} & \textbf{88.5} & \textbf{89.6} & \textbf{88.1} & \underline{54.1} & \underline{73.7} & \textbf{74.4} & \textbf{86.6} \\
        \bottomrule
    \end{tabular*}
    }
    \vspace{-0.5cm}
\end{table}

To assess multi-party spoken and image-text memory, we evaluate VoxPolyMem on VoxPolyBench and the public Mem-Gallery and H2HMem-Multi benchmarks. VoxPolyMem leads all three, with a weighted average of 86.6 versus 68.1 for the strongest external baseline (Table~\ref{tab:voxpolybench-main}).

On VoxPolyBench, VoxPolyMem achieves an overall score of 85.0, exceeding the strongest external baseline by 23.6 points, and leads all four task subsets. Personalized answering and interaction attribution show particularly strong results, scoring 76.1 and 87.5, respectively. These gains are consistent with interaction-aware memory that preserves speakers, addressees, and participant attributes, allowing retrieval to filter memories by participant roles and interaction relations. Its advantages also extend to retrieval and reasoning and memory evolution, suggesting that interaction context supports broader memory use beyond explicit personalization and attribution questions.

On the public Mem-Gallery and H2HMem-Multi benchmarks, VoxPolyMem achieves overall scores of 89.6 and 74.4, exceeding the strongest external baselines by 11.8 and 8.4 points, respectively. It leads all three Mem-Gallery task subsets and memory application on H2HMem-Multi, supporting the applicability of hierarchical memory and adaptive retrieval to general multimodal memory tasks. Notably, the w/o RL variant already exceeds all external baselines on each benchmark, achieving a weighted average of 84.4. This indicates that the memory structure and agentic retrieval process are effective without additional RL, while policy optimization provides further overall improvements.

\subsection{Component Ablations}
\label{sec:ablations}

\begin{table}[!t]
    \centering
    \small
    \setlength{\tabcolsep}{3pt}
    \renewcommand{\arraystretch}{1.12}
    \caption{Component ablations across three benchmarks. Ablated variants use w/o RL as the base model, with VoxPolyMem (full) included as a reference. Bold and underlined values mark the best and second-best results.}
    \label{tab:component-ablation}
    \begin{tabular*}{\linewidth}{@{\extracolsep{\fill}}lrrrrr@{}}
        \toprule
        & \multicolumn{2}{c}{\textbf{VoxPolyBench}} & \multicolumn{2}{c}{\textbf{Mem-Gallery}} & \textbf{H2HMem-Multi} \\
        \cmidrule(lr){2-3}\cmidrule(lr){4-5}\cmidrule(lr){6-6}
        Variant & Score & Recall & Score & Recall & Score \\
        \midrule
        \textbf{VoxPolyMem (full)} & \textbf{85.0} & \textbf{87.4} & \textbf{89.6} & \textbf{93.6} & \textbf{74.4} \\
        w/o RL & \underline{84.0} & \underline{85.2} & \underline{86.3} & 88.1 & \underline{72.1} \\
        \midrule
        w/o hierarchical memory & 79.8 & 81.4 & 84.9 & \underline{88.9} & 63.6 \\
        w/o interaction relations & 78.6 & 81.2 & 85.1 & 88.0 & 67.3 \\
        Single-round retrieval & 76.1 & 74.8 & 85.5 & 87.7 & 60.0 \\
        w/o query rewriting & 83.2 & 85.0 & 86.0 & 88.5 & 70.8 \\
        \bottomrule
    \end{tabular*}
    \vspace{-0.5cm}
\end{table}

Table~\ref{tab:component-ablation} evaluates memory construction and retrieval components using w/o RL as the base model, with full VoxPolyMem as a reference.

Removing fact memory and participant profiles restricts retrieval to original interaction records and lowers scores across all benchmarks, most on H2HMem-Multi, from 72.1 to 63.6. Removing interaction edges and relation-based filtering while retaining speaker identities affects VoxPolyBench most, reducing its score from 84.0 to 78.6. These patterns suggest that memory abstraction supports access to dispersed information, while explicit interaction structure helps distinguish evidence associated with different participants.

Single-round retrieval causes larger declines than retaining multiple rounds with the original query, indicating greater sensitivity to iterative retrieval than to query rewriting without RL. Follow-up searches allow the agent to gather evidence missing from earlier results, even when the query remains unchanged. RL improves scores from 86.3 to 89.6 on Mem-Gallery and from 72.1 to 74.4 on H2HMem-Multi. These gains support optimizing the joint retrieval policy to coordinate memory-layer selection, tool use, and query rewriting based on accumulated evidence.

\subsection{Comparison of Retrieval Policy Optimization Methods}
\label{sec:retrieval-policy-comparison}

Table~\ref{tab:retrieval-policy-comparison} compares reward design and credit assignment under fixed memory, retrieval tools, and inference budget. We adapt Search-R1~\citep{jin2025searchr1} and MoT-GRPO from Mem-T~\citep{yue2026memt} to our retrieval framework. Search-R1 uses a terminal answer reward, while MoT-GRPO combines tree-structured rollouts with evidence coverage and answer rewards. Terminal-coverage GRPO rewards final cumulative evidence coverage with a shared trajectory-level advantage. EG-GRPO instead assigns state-specific rewards and advantages for the coverage and ranking of newly acquired supporting evidence retained after each action.

\begin{table}[htbp]
    \centering
    \setlength{\abovecaptionskip}{3pt}
    \small
    \setlength{\tabcolsep}{2pt}
    \renewcommand{\arraystretch}{1.12}
    \caption{Comparison of retrieval policy optimization methods. Rounds denotes the mean number of executed retrieval rounds per question, excluding answer generation. Bold and underlined values mark the best and second-best results.}
    \label{tab:retrieval-policy-comparison}
    \begin{tabular*}{\linewidth}{@{\extracolsep{\fill}}lrrrrrrrr@{}}
        \toprule
        & \multicolumn{3}{c}{\textbf{VoxPolyBench}} & \multicolumn{3}{c}{\textbf{Mem-Gallery}} & \multicolumn{2}{c}{\textbf{H2HMem-Multi}} \\
        \cmidrule(lr){2-4}\cmidrule(lr){5-7}\cmidrule(lr){8-9}
        Method & Score $\uparrow$ & Recall $\uparrow$ & Rounds $\downarrow$ & Score $\uparrow$ & Recall $\uparrow$ & Rounds $\downarrow$ & Score $\uparrow$ & Rounds $\downarrow$ \\
        \midrule
        w/o RL & 84.0 & 85.2 & 1.6 & 86.3 & 88.1 & 1.7 & \underline{72.1} & 1.4 \\
        Search-R1 & 81.3 & 82.3 & 1.6 & 86.1 & 90.1 & 1.6 & 71.9 & \underline{1.3} \\
        Terminal-coverage GRPO & 83.1 & \underline{86.7} & \underline{1.4} & 86.4 & 89.7 & \underline{1.5} & 71.6 & 1.5 \\
        MoT-GRPO & \underline{84.1} & 86.6 & \underline{1.4} & \underline{87.1} & \underline{91.5} & 1.8 & 72.0 & 1.6 \\
        \textbf{EG-GRPO (ours)} & \textbf{85.0} & \textbf{87.4} & \textbf{1.2} & \textbf{89.6} & \textbf{93.6} & \textbf{1.3} & \textbf{74.4} & \textbf{1.2} \\
        \bottomrule
    \end{tabular*}
    \vspace{-5pt}
\end{table}

\FloatBarrier
Compared with w/o RL, Search-R1 lowers answer scores across all three benchmarks. Terminal-coverage GRPO improves recall on both annotated benchmarks without consistent answer improvements, showing that higher evidence coverage does not necessarily yield better answers. MoT-GRPO improves both metrics on VoxPolyBench and Mem-Gallery. EG-GRPO leads all three benchmarks in answer scores and both annotated benchmarks in recall. On Mem-Gallery, it raises answer scores from 87.1 to 89.6 and recall from 91.5 to 93.6 compared with MoT-GRPO.

These improvements accompany fewer retrieval rounds. EG-GRPO uses an average of 1.2--1.3 rounds across benchmarks, compared with 1.4--1.7 for w/o RL and 1.4--1.8 for MoT-GRPO. Its coverage reward encourages complementary searches, while its ranking reward favors newly acquired supporting evidence near the top of the selected context. These results support the overall training design for improving evidence acquisition and answering with fewer retrieval rounds.

\section{Conclusion}

To support persistent memory of participant identities and interactions across spoken conversations, we propose VoxPolyMem, combining incremental speaker identification, interaction-aware hierarchical memory, and adaptive retrieval. EG-GRPO trains the retrieval policy using round-wise evidence gains. We also introduce VoxPolyBench to evaluate memory over multi-session spoken interactions. Experiments on VoxPolyBench and two public image-text benchmarks support the effectiveness of the memory structure and agentic retrieval, with further improvements from policy optimization. Future work will extend evaluation to real-world spoken interactions with greater acoustic and conversational diversity. Further limitations appear in Appendix~\ref{app:limitations}.

\subsection*{AI Use Statement}
LLMs were used for benchmark construction and the memory and evaluation pipelines, as described in Sections~\ref{sec:benchmark}--\ref{sec:experiments}. Generative AI tools also assisted code development. The authors take responsibility for the final manuscript and research artifacts.

\subsection*{Ethics Statement}
VoxPolyBench uses generated dialogues and synthesized speech. This controlled setting does not establish robustness across real speakers, accents, or recording conditions. Voiceprint embeddings, identity links, and participant profiles can contain sensitive information and enable unwanted tracking or disclosure. Real-world deployment should require informed consent, restricted access, limited retention, and mechanisms to correct or delete stored information. Benchmark performance should not be interpreted as authorization for surveillance or high-stakes identity decisions.

\subsection*{Reproducibility Statement}
Sections~\ref{sec:benchmark} and~\ref{sec:method} describe benchmark construction and the method. Appendices~\ref{app:dataset-statistics}--\ref{app:additional-experiments} provide data validation, implementation settings, evaluation protocols, and additional experiments. Reproducibility with hosted models also depends on API versions and sampling variability.

\bibliography{iclr2027_conference}
\bibliographystyle{iclr2027_conference}

\appendix
\section{Benchmark Details}
\label{app:dataset-statistics}
\label{app:benchmark-composition}

\subsection{Construction and Quality Control}
\label{app:quality-control}

Quality control covers session anchors and the resulting QA annotations. Before dialogue generation, human reviewers inspect each session's anchors for consistency with participant settings and the planned event progression. After generation, QA pairs undergo consistency checks by both human reviewers and an LLM.

The checks address four requirements. Realized events must agree with the anchors and their temporal updates. Speaker and addressee assignments must agree with the dialogue. Questions must be clear, and reference answers must be supported by the conversation. Annotated evidence must be sufficient to support the answer. Correctable errors are revised; QA pairs with unresolved ambiguity or insufficient support are removed. These checks concern dataset construction, separately from the human audit of model-answer scores in Appendix~\ref{app:human-llm-judge-agreement}.

\paragraph{Speech synthesis quality control.}
We assign each participant a fixed reference voice across sessions and convert telephone numbers and alphanumeric identifiers to character-by-character pronunciations. Synthesis uses short reference clips and a target duration estimated from text length. We generate up to three candidates, preferring one within the duration tolerance and otherwise retaining the candidate with the smallest duration deviation. Duration serves as an auxiliary screen for potential abnormal lengthening or truncation. Post-synthesis ASR checks flag normalized text similarity below 0.85, telephone-number or identifier mismatches, and suspected repetitions. During construction, selected problematic samples undergo targeted resynthesis and transcription rechecks, supplemented by manual spot listening.

\paragraph{Speech transcription consistency.}
\label{app:speech-consistency}
We independently transcribe all 9,599 final dialogue audio segments across the 18 scenarios using Whisper large-v3-turbo, with English recognition, temperature 0, and beam size 5. No reference texts or participant-name hints are supplied to the recognizer. Each reference is the complete source text used to synthesize the corresponding audio segment. We apply identical normalization to references and ASR outputs: curly apostrophes are mapped to straight apostrophes, em dashes to spaces, and non-breaking hyphens to standard hyphens, followed by Whisper's EnglishTextNormalizer to normalize case, punctuation, abbreviations, and number expressions. We do not correct transcriptions using the references.

We compute word-level edit distances for individual segments and aggregate substitution, deletion, and insertion counts over the corpus:
\begin{equation}
\mathrm{WER}=\frac{\sum_{i=1}^{M}(S_i+D_i+I_i)}{\sum_{i=1}^{M}N_i},
\end{equation}
where $M$ is the number of evaluated segments, $S_i$, $D_i$, and $I_i$ are the respective edit counts, and $N_i$ is the number of normalized reference words in segment $i$. The resulting corpus-level WER is 2.14\%; it is weighted by reference word count rather than averaged equally across segments. This metric measures agreement between automatic transcriptions and synthesis source texts, rather than a human-verified synthesis error rate.

\subsection{Dataset Statistics}
\label{app:dataset-composition}

\begin{figure}[!t]
    \centering
    \begin{minipage}[b]{0.54\linewidth}
        \centering
        \includegraphics[width=\linewidth]{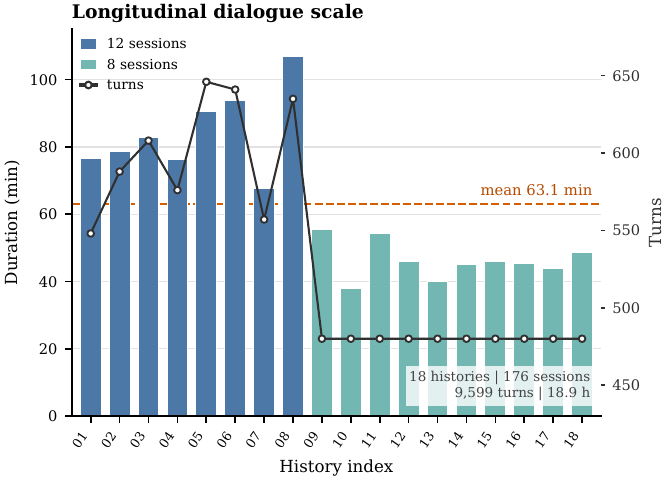}
        \par\smallskip\footnotesize\textbf{(a)} Longitudinal dialogue scale.
    \end{minipage}
    \hfill
    \begin{minipage}[b]{0.43\linewidth}
        \centering
        \includegraphics[width=\linewidth]{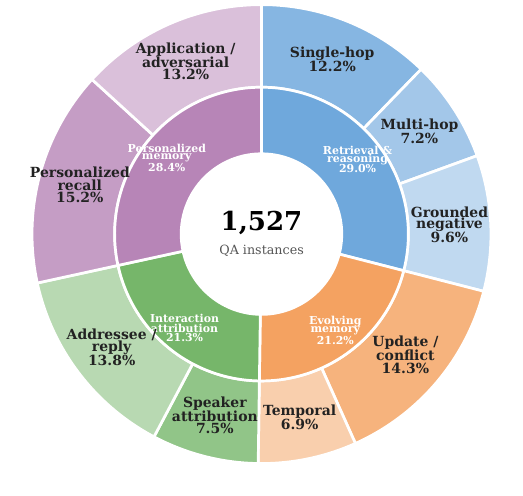}
        \par\smallskip\footnotesize\textbf{(b)} Hierarchical memory-task taxonomy.
    \end{minipage}

    \vspace{2mm}
    \begin{minipage}[b]{\linewidth}
        \centering
        \includegraphics[width=\linewidth]{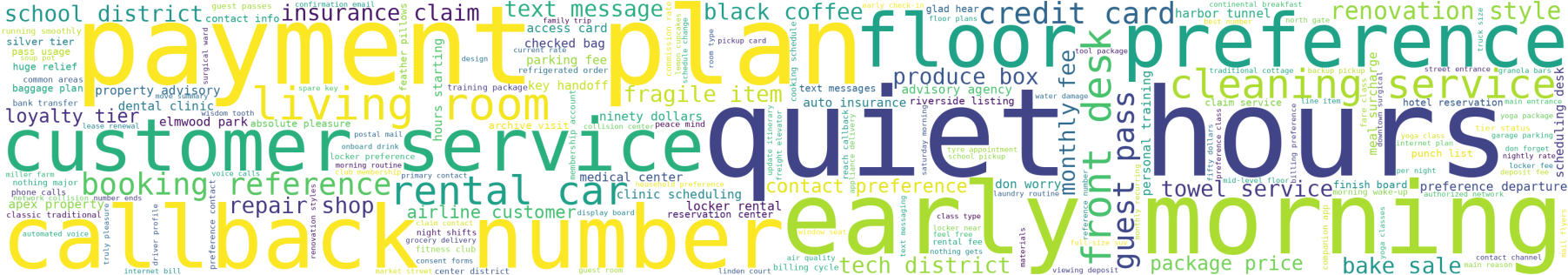}
        \par\smallskip\footnotesize\textbf{(c)} Dialogue-domain coverage across 9,599 turns.
    \end{minipage}
    \caption{VoxPolyBench dataset statistics. (a) Duration and turn count for 18 case-level histories, comprising 176 sessions and 18.9 hours of dialogue audio; cases are identified only by indices 01--18. (b) Distribution of the 1,527 QA instances in VoxPolyBench: the inner ring shows four coarse task families and the outer ring shows nine fine-grained types. (c) Case-balanced content terms extracted exclusively from dialogue turns after removing participant names, identifiers, quantities, and conversational filler.}
    \label{fig:voxpolybench-dataset-statistics}
\end{figure}

\paragraph{Longitudinal dialogue structure.}
VoxPolyBench contains 18 case-level histories spanning 176 sessions, 9,599 dialogue turns, and 18.9 hours of synthesized speech. Each case follows a recurring group of participants over 8--12 sessions and averages 63.1 minutes of dialogue, while the average session lasts 6.5 minutes. The resulting evaluation unit is a longitudinal multi-party history rather than an isolated recording or a collection of independent utterances. Figure~\ref{fig:voxpolybench-dataset-statistics}(a) reports the case-level duration and turn distribution, and Figure~\ref{fig:voxpolybench-dataset-statistics}(c) summarizes domain coverage using case-balanced dialogue terms.

\paragraph{Memory-task composition.}
VoxPolyBench contains 1,527 QA instances organized under a hierarchical taxonomy. Retrieval and reasoning accounts for 443 instances (29.0\%), evolving memory for 324 (21.2\%), interaction attribution for 326 (21.3\%), and personalized memory for 434 (28.4\%). The nine fine-grained task types cover single- and multi-hop retrieval, grounded negatives, memory updates and conflicts, temporal reasoning, source-speaker attribution, addressee attribution, personalized recall, and application or adversarial questions. This distribution preserves conventional retrieval and temporal tasks while assigning substantial coverage to participant-dependent memory, as visualized in Figure~\ref{fig:voxpolybench-dataset-statistics}(b).

\subsection{Comparison with Existing Benchmarks}
\label{app:benchmark-comparison}
Table~\ref{tab:benchmark-capability-comparison} separates three properties that are often conflated: spoken-dialogue understanding, multi-user personalization, and long-term multi-party memory. Existing speech benchmarks primarily evaluate within-session understanding or long-context recall without persistent participant identity. AFA routes independently interacting users to isolated persona memories, but does not represent a shared multi-party dialogue or evaluate structured long-horizon memory reasoning. Conversely, multimodal memory benchmarks provide longitudinal histories and richer memory operations, but generally expose participant labels and do not jointly evaluate spoken identity, source-speaker attribution, addressee attribution, and query-participant conditioning. VoxPolyBench combines these properties with memory evolution, cross-session evidence integration, and gold evidence provenance. To our knowledge, it is the first benchmark for long-term memory reasoning over multi-session, multi-party spoken conversations.

The comparison draws on ContextDialog~\citep{kim2025does}, MSU-Bench~\citep{wang2025msubench}, AFA~\citep{alratrout2026afa}, the unpublished Vox-Infinity manuscript~\citep{xu2026voxinfinity}, Mem-Gallery~\citep{bei2026memgallery}, and H2HMem~\citep{zhu2026h2hmem}. Interaction attribution receives full coverage only when both source-speaker and addressee attribution are evaluated; coverage of only one is marked partial. The table summarizes targeted capabilities rather than ranking overall benchmark quality. Gold supporting evidence and cross-session evidence integration remain relevant evaluation properties, discussed here rather than shown as separate columns.

\section{Implementation Details}
\label{app:implementation}

\subsection{Acoustic Speaker Tracking}
\label{app:speaker-identification}

\noindent\textbf{Online configuration.}
The matching and EMA-update thresholds are $\tau_{\mathrm{match}}=0.35$ and $\tau_{\mathrm{update}}=0.40$, with update weight $\alpha=0.05$. Per-turn voiceprint embeddings and assignments are retained for subsequent identity reconciliation.

\noindent\textbf{Candidate comparison.}
After each session and at the end of the input stream, utterances observed so far are grouped by their original speaker identifiers. Candidate pairs are scored using cross-group cosine similarities. Only a group at most one-third the size of its target is eligible to merge. A multi-utterance group's score is the median cross-group similarity. A singleton's score is its mean similarity to the five closest utterances in the target group.

\noindent\textbf{Merging criteria.}
Multi-utterance groups must be mutual best matches, with sufficient separation from alternative candidates. Their cross-group similarity is assessed relative to within-group cohesion, and they must share at least half of the sessions represented by the group with fewer sessions. Singletons require a target containing at least ten utterances, a similarity score of at least $0.35$, and sufficient separation from the runner-up when one exists.

The highest-scoring eligible pair is merged, candidates are recomputed, and the process stops when no eligible pair remains. The resulting alias mapping preserves original voiceprint embeddings and online EMA entries.

\subsection{Memory and Retrieval Configuration}
\label{app:memory-retrieval-settings}

Acoustic reconciliation and context-aware name association produce identity mappings while preserving the underlying voiceprint embeddings. The former links fragmented speaker identifiers, while the latter associates them with participant names and profiles. Filters are specified through action parameters or resolved from the asker identity. Empty filters impose no participant constraints, and automatically supplied identity conditions are excluded from the policy loss. The final-context limit of $k=15$ applies after RRF fusion, deduplication, and context selection, rather than separately to each tool call.

\subsection{Evidence-Gain GRPO Details}
\label{app:eg-grpo}

\paragraph{Source evidence and ranking.}
The budget $k=15$ limits retained memory records after RRF fusion and deduplication. These records, including extracted facts and profiles, form the answer context $C_i$. Source-ID mapping is used for reward calculation and does not replace these memories with source-turn text. We obtain $U_i$ from retained interaction-turn IDs and fact/profile \texttt{refer\_id} links, without an additional source-ID truncation. Multiple source IDs linked to one memory share its rank. If an ID occurs in several memories, its earliest memory rank defines $p_i(e)$, and it contributes only once. Sibling candidates are evaluated separately.

The denominators $|G_q|$ and $Z_q$ provide fixed scales for a question. Because several supporting IDs may share one memory rank, $N_i$ is a rank-weighted evidence gain rather than standard NDCG, and is not necessarily bounded by one. Candidates share the pre-action history and denominators, and their advantages are standardized within the current state. Empty retained contexts receive zero reward. Training requires nonempty supporting-turn annotations.

\paragraph{Branch history and expansion.}
The successor history is $B_i^{\mathrm{next}}=B\cup U_i$, recording source IDs represented in retained memories. A source ID returned only through discarded memories is not added to this history and remains eligible for reward when subsequently retained. Siblings share the pre-action history but never contribute to each other's histories.

Each successor reduces the round budget by one. Training expansion stops when the current retained memories cover all supporting IDs, $G_q\subseteq U_i$, or the budget is exhausted. Historical coverage alone does not stop a branch. The completing action remains in its current training group. At inference, stopping uses evidence sufficiency and the round budget without gold annotations.

\paragraph{Policy optimization.}
The state-specific advantages in \eqref{eq:evidence-objective} enter the objective in \eqref{eq:grpo-objective}. Let $a_{i,n}$ be token $n$ of action $i$, and $a_{i,<n}$ its preceding tokens. The probability ratio is $w_{i,n}=\pi_\theta(a_{i,n}\mid s_t,a_{i,<n})/\pi_{\mathrm{old}}(a_{i,n}\mid s_t,a_{i,<n})$, and $\mathrm{KL}_{i,n}=D_{\mathrm{KL}}[\pi_\theta(\cdot\mid s_t,a_{i,<n})\|\pi_{\mathrm{ref}}(\cdot\mid s_t,a_{i,<n})]$. The reference policy $\pi_{\mathrm{ref}}$ is fixed. Automatically supplied filters and all input or retrieved tokens are excluded from the action loss. The expectation in \eqref{eq:grpo-objective} covers sampled states and candidate groups. The evidence reward supervises retrieval actions, while the answer model remains fixed.

\paragraph{Example.}
Suppose $G_q=\{g_1,g_2\}$ and $B=\{g_1\}$. If three retained memory records support $g_1$, neither gold ID, and $g_2$, respectively, then $D_i=\{g_2\}$ and $p_i(g_2)=3$. Coverage is $1/2$, and $N_i=(1/\log_2 4)/(1+1/\log_2 3)\approx0.307$. If the memory supporting $g_2$ is discarded, $g_2$ is absent from $U_i$ and remains eligible for reward in a later round.

\section{Evaluation Protocols}
\label{app:evaluation-details}

\subsection{Data Splits and Baselines}
\label{app:comparison-settings}

Table~\ref{tab:train-test-counts} provides category-level QA counts for the three evaluation benchmarks. The training sources and shared evaluation settings are described in Section~\ref{sec:experimental-setup}.
\begin{table}[htbp]
    \centering
    \small
    \setlength{\tabcolsep}{8pt}
    \renewcommand{\arraystretch}{1.05}
    \caption{Training and test QA counts. Total includes both splits; H2HMem-Multi is used only for testing. Category abbreviations follow Table~\ref{tab:voxpolybench-main}; test counts determine the weights for score aggregation.}
    \label{tab:train-test-counts}
    \begin{tabular}{llrrr}
        \toprule
        Benchmark & Category & Total & Train & Test \\
        \midrule
        VoxPolyBench & RR & 443 & 60 & 383 \\
         & EC & 324 & 40 & 284 \\
         & P & 434 & 60 & 374 \\
         & IA & 326 & 40 & 286 \\
         & \textbf{Total} & \textbf{1,527} & \textbf{200} & \textbf{1,327} \\
        \midrule
        Mem-Gallery & E\&A & 862 & 167 & 695 \\
         & R & 503 & 103 & 400 \\
         & KM & 346 & 30 & 316 \\
         & \textbf{Total} & \textbf{1,711} & \textbf{300} & \textbf{1,411} \\
        \midrule
        H2HMem-Multi & A & 75 & 0 & 75 \\
         & R & 49 & 0 & 49 \\
         & Rec & 51 & 0 & 51 \\
         & \textbf{Total} & \textbf{175} & \textbf{0} & \textbf{175} \\
        \bottomrule
    \end{tabular}
\end{table}

\paragraph{Baselines.}
The text-based memory baselines are mem0~\citep{chhikara2025mem0}, A-Mem~\citep{xu2025amem}, and LightMem~\citep{fang2025lightmem}. The multimodal retrieval and memory baselines are AUGUSTUS~\citep{jain2025augustus}, UniversalRAG~\citep{yeo2026universalrag}, M2A~\citep{feng2026m2a}, MemVerse~\citep{liu2025memverse}, NGMemory~\citep{fisher2025ngm}, HippoMM~\citep{lin2025hippomm}, and M3-Agent~\citep{long2025seeing}.

\subsection{Metrics and Aggregation}
\label{app:metrics}

\paragraph{Score aggregation.}
Within each benchmark, category scores are weighted by their test-question counts. The cross-benchmark Average uses the total test counts in Table~\ref{tab:train-test-counts} as weights. Aggregates, differences, rankings, and relative reductions are computed before display rounding. A dash denotes an unavailable measurement.

\paragraph{Task grouping.}
Temporal reasoning belongs to EC in VoxPolyBench, following Figure~\ref{fig:voxpolybench-dataset-statistics}. Mem-Gallery's nine tasks are grouped into three dimensions and aggregated by QA count. E\&A comprises Factual Retrieval, Visual-centric Search, and Test-Time Learning. R comprises Temporal Reasoning, Visual-centric Reasoning, and Multi-entity Reasoning. KM comprises Knowledge Resolution, Conflict Detection, and Answer Refusal. The dimension names and abbreviations follow Table~\ref{tab:voxpolybench-main}.

\paragraph{Evidence recall.}
Recall is measured after final-context deduplication and truncation. Retained facts and profiles contribute the source turn IDs linked through their \texttt{refer\_id} fields, with each source turn ID counted once. Gold supporting-turn annotations are held fixed across variants. The per-question definition and averaging rule are given in Section~\ref{sec:experimental-setup}.

\subsection{LLM-as-a-Judge Protocol}
\label{app:judge-protocol}

The judges specified in Section~\ref{sec:experimental-setup} independently receive the same prompt below. Each request consists of a single user message containing the question, reference answer, and system answer.

\noindent\textbf{Judge prompt template.}
\begin{Verbatim}[fontsize=\footnotesize,breaklines=true,breakanywhere=true]
You are an impartial judge evaluating the memory capabilities of an AI assistant with the question-answering task.
Your task is to compare the Assistant's Answer against the Ground Truth and assign a score of 0, 0.25, 0.5, 0.75, or 1.

### Scoring Rubric

**Score 0 (Incorrect / Miss):**

- The answer contradicts the Ground Truth.
- For Yes/No questions: The answer has the wrong polarity (e.g., says "Yes" when Ground Truth is "No").
- For Open-ended questions: The answer provides factually wrong information or hallucinations.
- The assistant fails to provide the required information.

**Score 0.25 (Poor / Tangential):**

- The answer touches on the topic but misses the **core entity** or key value required.
- The answer contains a mix of minor correct details and **significant hallucinations** or wrong associations.
- The answer is excessively vague to the point of being useless (e.g., answering "a dog" instead of "a golden retriever").

**Score 0.5 (Partial / Vague):**

- The answer is technically correct, but lacks confidence or is incomplete.
- The answer captures the **main entity or concept** correctly but misses a part of the required supporting details.
- For Yes/No questions: The polarity is correct, but the reasoning is flawed (if have), or the assistant is uncertain (e.g., "I think it might be Yes").
- For Open-ended questions: The answer is too general or misses key adjectives/details present in the Ground Truth.

**Score 0.75 (Good / Minor Imperfection):**

- The answer is largely accurate and captures the core information confidently.
- It misses only **minor details** (e.g., specific adjectives or secondary details) that do not alter the main truth.
- The answer contains all the correct information but includes unnecessary "fluff" or slight conversational filler that reduces precision.

**Score 1 (Correct / Exact):**

- The answer is accurate, precise, and confident.
- For Yes/No questions: The polarity matches the Ground Truth perfectly.
- For Open-ended questions: The answer contains **all** the core information and necessary details required by the Ground Truth without hallucinations.

### Input Data

Question: {question}
Ground Truth: {ground_truth}
Assistant Answer: {model_output}

### Output Format

Output strictly in the following JSON format:
{"score": <0, 0.25, 0.5, 0.75, or 1>, "reasoning": "<short explanation>"}
\end{Verbatim}

\section{Additional Experiments}
\label{app:additional-experiments}

\subsection{Speaker Identity Evaluation across Three Datasets}
\label{app:speaker-ablation}

\paragraph{Protocol.}
We evaluate anonymous speaker-ID tracking on the 18 VoxPolyBench cases
(9,599 turns), IEMOCAP Sessions 4--5 (4,273 utterances), and five AMI
evaluation scenario chains (1,311 non-overlapping utterances). All variants
reuse frozen ECAPA embeddings and oracle utterance boundaries, so the study
isolates identity tracking rather than end-to-end diarization. Reference
speaker labels and speaker counts are used only for scoring; inference receives
neither a participant roster nor a target number of speakers.

\paragraph{Component analysis.}
Table~\ref{tab:speaker-online-ablation} compares the static tracker with cumulative additions of confidence-gated EMA updates and identity reconciliation across all three datasets. EMA updates improve speaker attribution accuracy on all datasets, while identity reconciliation provides further gains, most notably on IEMOCAP, from 85.0\% to 95.0\%.

\begin{table}[htbp]
    \centering
    \small
    \setlength{\tabcolsep}{12pt}
    \caption{Cumulative ablation of the shared anonymous speaker tracker.
    Each cell reports speaker attribution accuracy (\%) after optimal
    permutation matching, with macro ARI $\times 100$ in parentheses. Neither reference
    identities nor the number of speakers is available during inference.}
    \label{tab:speaker-online-ablation}
    \begin{tabular}{lccc}
        \toprule
        Variant & VoxPolyBench $\uparrow$ & IEMOCAP $\uparrow$ & AMI $\uparrow$ \\
        \midrule
        Static prototype
          & 90.3 (90.0) & 57.4 (35.7) & 70.3 (58.6) \\
        $+$ online EMA
          & 97.5 (98.1) & 85.0 (71.4) & 87.3 (81.4) \\
        $+$ identity reconciliation
          & \textbf{97.7 (98.4)} & \textbf{95.0 (80.8)} & \textbf{87.9 (82.4)} \\
        \bottomrule
    \end{tabular}
\end{table}

\subsection{Human Audit of LLM-Judge Scores}
\label{app:human-llm-judge-agreement}

We compare aggregate LLM-judge and human scores on VoxPolyBench and Mem-Gallery. We sample 100 responses from each benchmark's complete w/o RL evaluation output using the fixed seed 20260920. Sampling preserves the judge-score distribution in proportion to the full result pool and distributes the selected items across the native task categories. Table~\ref{tab:human-llm-judge-agreement} reports the mean LLM-judge score, the mean human score, and their signed and absolute gaps. All scores are measured on the same five-level rubric and multiplied by 100.

\begin{table}[htbp]
    \centering
    \small
    \setlength{\tabcolsep}{3pt}
    \renewcommand{\arraystretch}{1.12}
    \caption{Human audit of LLM-judge scores. $\Delta$ is Human $-$ LLM, and $|\Delta|$ is the absolute mean-score gap.}
    \label{tab:human-llm-judge-agreement}
    \begin{tabular*}{\linewidth}{@{\extracolsep{\fill}}lrrrrr@{}}
        \toprule
        Benchmark & $N$ & LLM & Human & $\Delta$ & $|\Delta|$ \\
        \midrule
        VoxPolyBench & 100 & 83.3 & 85.0 & +1.8 & 1.8 \\
        Mem-Gallery & 100 & 86.0 & 87.3 & +1.3 & 1.3 \\
        \midrule
        Overall & 200 & 84.6 & 86.1 & +1.5 & 1.5 \\
        \bottomrule
    \end{tabular*}
\end{table}

The sampled LLM means differ from the w/o RL test scores of 84.0 on VoxPolyBench and 86.3 on Mem-Gallery by 0.8 and 0.3 points, respectively. Human means are slightly higher on both benchmarks, while the absolute gaps remain within 1.8 points. Across the 200 sampled responses, the aggregate LLM and human means are 84.6 and 86.1, respectively. Similar means do not establish item-level agreement between human and automatic scores.

\section{Limitations}
\label{app:limitations}

VoxPolyBench uses generated dialogues and synthesized speech, with human review of event plans and QA pairs. Constructing challenging multi-hop questions from existing long conversations requires reliable links between dispersed evidence and unambiguous answers, which are difficult to ensure through automatic generation alone. Our current construction therefore uses planned events and human validation to control question and evidence quality. Future work will explore question generation and validation over transcripts of real long conversations, paired with their original audio, to broaden coverage of complex reasoning and real-world spoken interactions.

\end{document}